\documentclass[10pt,twocolumn,letterpaper]{article}

\usepackage{cvpr}              

\usepackage{graphicx}
\usepackage{amsmath}
\usepackage{amssymb}
\usepackage{booktabs}
\usepackage{stfloats}
\usepackage{amsfonts}       
\usepackage{nicefrac}       
\usepackage{microtype}      
\usepackage{xcolor}         
\usepackage{graphicx}
\usepackage{diagbox}
\usepackage{multicol}
\usepackage{enumerate}
\usepackage{times}
\usepackage{epsfig}
\usepackage{graphicx}
\usepackage{amsmath}
\usepackage{amssymb}
\usepackage{threeparttable}
\usepackage{enumitem}
\usepackage{multirow}
\usepackage{color}
\usepackage{array}

\usepackage{ulem}

\usepackage[colorlinks,linkcolor=red]{hyperref}
\usepackage[numbers,sort&compress]{natbib}
\setenumerate[1]{leftmargin = 10pt, itemsep=0pt,partopsep=0pt,parsep=-0pt,topsep=-0pt}
\usepackage{hyperref}

\newcommand{\PreserveBackslash}[1]{\let\temp=\\#1\let\\=\temp}

\usepackage[capitalize]{cleveref}
\crefname{section}{Sec.}{Secs.}
\Crefname{section}{Section}{Sections}
\Crefname{table}{Table}{Tables}
\crefname{table}{Tab.}{Tabs.}

\def\cvprPaperID{5008} 
\def\confName{CVPR}
\def\confYear{2022}
\def\hlinew#1{%
 \noalign{\ifnum0=`}\fi\hrule \@height #1 \futurelet
 \reserved@a\@xhline}
 
\begin{document}


\title{1st Solution Places for CVPR 2023 UG$^{\textbf{2}}$+ Challenge Track 2.2-\\Coded Target Restoration through Atmospheric Turbulence}

\author{Shengqi Xu, Shuning Cao, Haoyue Liu, Xueyao Xiao, Yi Chang\footnotemark[1], Luxin Yan\\
National Key Lab of Multispectral Information Intelligent Processing Technology,\\
Huazhong University of Science and Technology, China\\
{\tt\small \{m202273123, sn\_cao, liuhy, xiaoxueyao, yichang, yanluxin\}@hust.edu.cn}}
\maketitle

\begin{abstract}
  In this technical report, we briefly introduce the solution of our team “VIELab-HUST” for coded target restoration through atmospheric turbulence in CVPR 2023 UG$^{2}$+ Track 2.2. In this task, we propose an efficient multi-stage framework to restore a high quality image from distorted frames. Specifically, each distorted frame is initially aligned using image registration to suppress geometric distortion. We subsequently select the sharpest set of registered frames by employing a frame selection approach based on image sharpness, and average them to produce an image that is largely free of geometric distortion, albeit with blurriness. A learning-based deblurring method is then applied to remove the residual blur in the averaged image. Finally, post-processing techniques are utilized to further enhance the quality of the output image. Our framework is capable of handling different kinds of coded target dataset provided in
  the final testing phase, and \textbf{ranked 1st on the final leaderboard}. Our code will be available at \url{{https://github.com/xsqhust/Turbulence_Removal}}.

\end{abstract}

\vspace{-10pt}
\section{Introduction}
This technical report presents our solutions to CVPR 2023 UG$^{2}$+ Challenge Track 2.2-Coded Target Restoration Through Atmospheric Turbulence. The participant's task is to improve the image quality so that the infomation encoded in the target patterns can be successfully decoded.

\begin{figure}[t]
  \vspace{0cm}  
 \setlength{\abovecaptionskip}{0.3 cm}   
 \setlength{\belowcaptionskip}{0 cm}   
   \centering
      \includegraphics[width=1.00\linewidth]{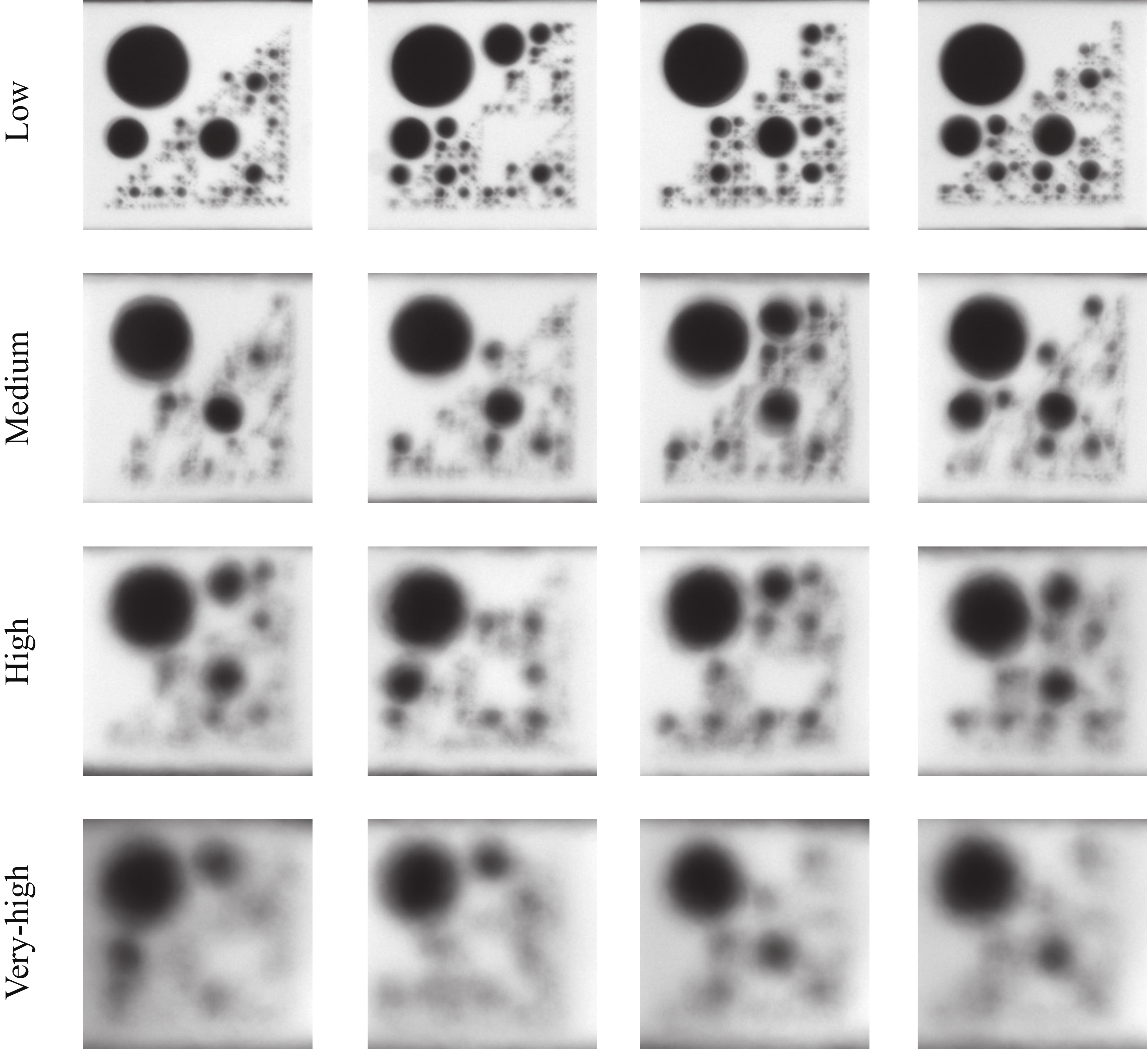}
   \caption{The visual examples of different degradation levels of code images.}
   \label{dataset}
 \end{figure}

As shown in Fig.\ref{dataset}, coded images with four levels of intensity are provided in the final testing phase: low, medium, high, very-high. Each level of turbulence degradation comprises 48 sequences, each consisting of 100 distorted frames. The evaluation metric for this phase is the average bit score calculated from the reconstruction result of the code patterns.

We propose a multi-stage framework to mitigate the effects caused by the turbulence. Firstly, each distorted frame is aligned using optical-flow based Image registration to suppress geometric distortion. Next, We select the sharpest set of registered frames
by employing a frame selection approach based on image
sharpness, and average them to produce an image that
is largely free of geometric distortion, albeit with blurriness. Then, a learning-based deblur method is applied to remove the blur in the average image.
Finally, post-processing is utilized to further enhance the quality of the output image. Our framework can handle different kinds of coded target dataset and ranked 1st on the final leaderboard. 
 
 The technical report is organized as follows: In Section \ref{sec2}, we provide a brief description of our restoration framework. Section \ref{sec3} provides an introduction to the training dataset. Section \ref{sec4} presents the experimental results, which demonstrate our framework's performance relative to other methods..
 \begin{figure*}[t]
  \centering
     \includegraphics[width=1.00\linewidth]{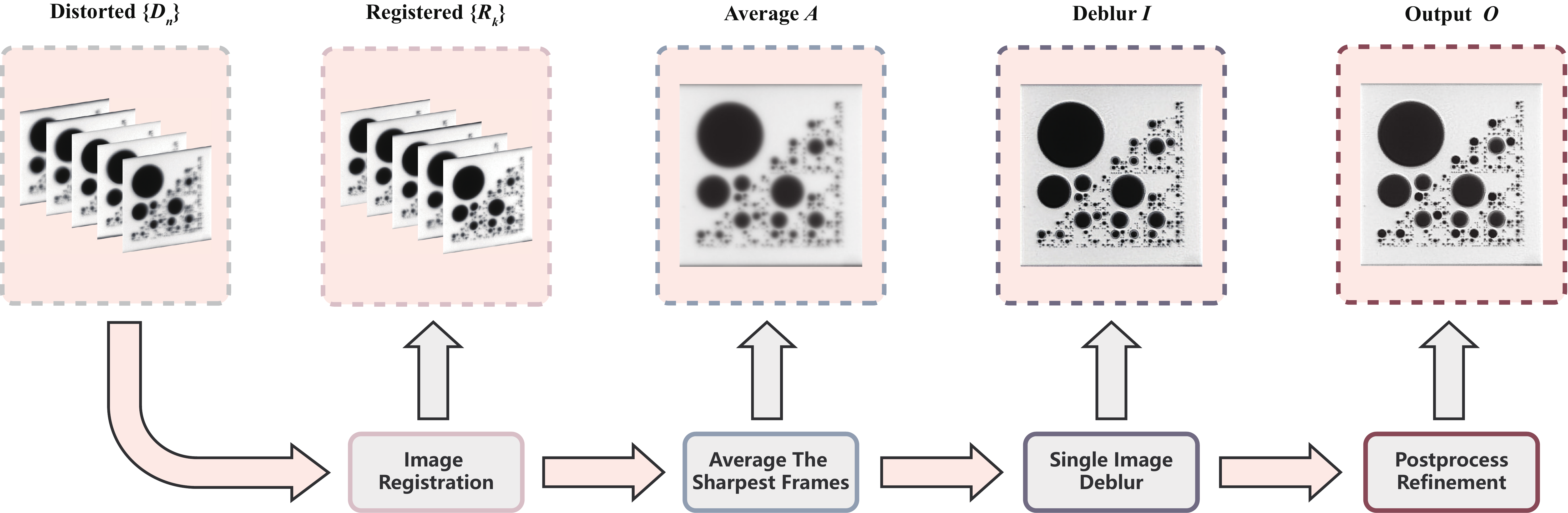}
  \caption{Block diagram for the proposed restoration framework. We firstly utilize
  image registration based on optical-flow to suppress geometric distortion. Next, we select the sharpest set of registered
  frames using frame selection based on sharpness and average them to produce an image approximately free of geometric distortion but with blurriness. Then, a learning-
  based deblur method is applied to remove the blur in the average image. Finally, post-processing is utilized to further enhance the quality of the output image.}
  \label{Framework}
\end{figure*}
\section{Restoration Framework}\label{sec2}
The proposed restoration framework contains four main steps (see the diagram in Fig. \ref{Framework}):

\textbf{A.} Image Registration;

\textbf{B.} Average The sharpest Frames;

\textbf{C.} Singe Image Deblur;

\textbf{D.} Postprocess Refinement.

\subsection{Image Registration}\label{sec2.1}
In step A, each frame in the distorted frames \textbf{\{\textit{D$_n$}\}} is aligned with a reference frame using optical flow, generating a registered sequence \textbf{\{\textit{R$_k$}\}} with less geometric distortion, 
and the reference frame is constructed by averaging the selected sequence \textbf{\{\textit{D$_n$}\}}. The
purpose of this step is to suppress geometric distortion.

\subsection{Average The Sharpest Frames}\label{sec2.2}
\begin{figure}[tb]
     \centering
        \includegraphics[width=1.00\linewidth]{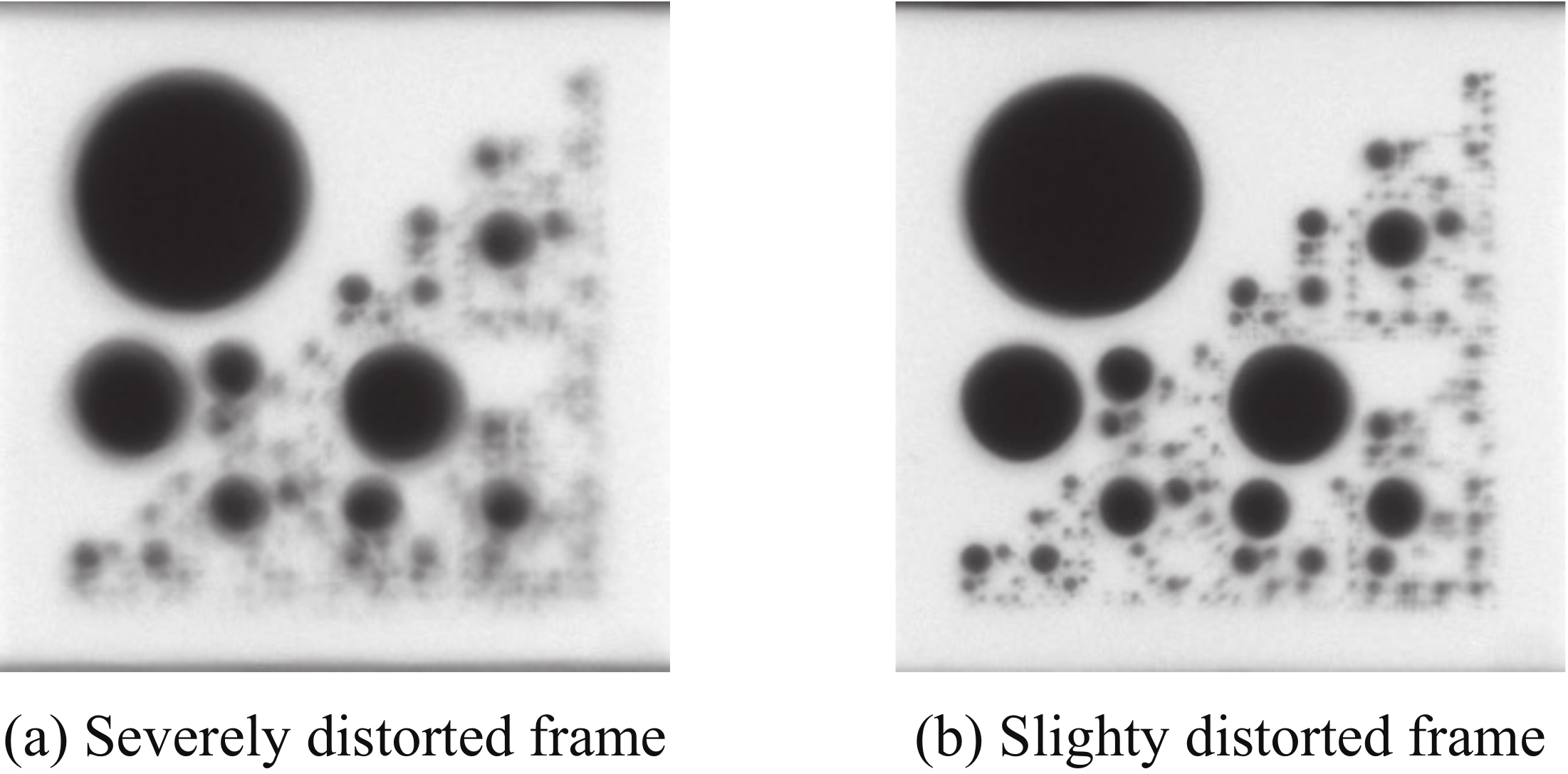}
     \caption{Frames with different visual quality in a sample sequence distorted
     by the atmospheric turbulence. (a) is a severely distorted frame that has a negative contribution in restoring the high-quality image, while (b) is a slightly distorted frame containing more useful information.}
     \label{diff}
\end{figure}

When imaging through a turbulent medium, atmospheric turbulence affects frames in a sequence unequally.The degree of distortion varies from frame to frame due to random fluctuations
of the refractive index in the optical transmission path. Consequently, some frames have better image quality than others,
with less blurriness and more useful image information. 
This point is illustrated in Fig. \ref{diff}, where two sample frames of a code sequence that is distorted by atmospheric turbulence are shown. Fig. \ref{diff}(a) is a frame that is severely distorted by the turbulence and Fig. \ref{diff}(b) is a sharp frame from the same sequence. By comparison, it can be found that Fig. \ref{diff}(a) has a negative contribution to the restoration of the
image.

Given an observed sequence \textbf{\{\textit{D$_n$}\}}, each frame in the sequence distorted by atmospheric turbulence can have different visual quality than the other frames. Sharpness is one of the most important image
quality factors since it determines the amount of detail information an image can convey. Therefore, A frame selection algorithm based on sharpness is firstly utilized to select the 
sharpest set of distorted frames that can help accurately reconstrut a high-quality image. In this step, the sharpness is computed by the intensity gradients of the image. The selected frames are then averaged to produce an image \textbf{\{\textit{A}\}} that approximately free of geometric distortion, albeit with some blurriness.

 \begin{figure*}[htpb]
  \centering
     \includegraphics[width=1.00\linewidth]{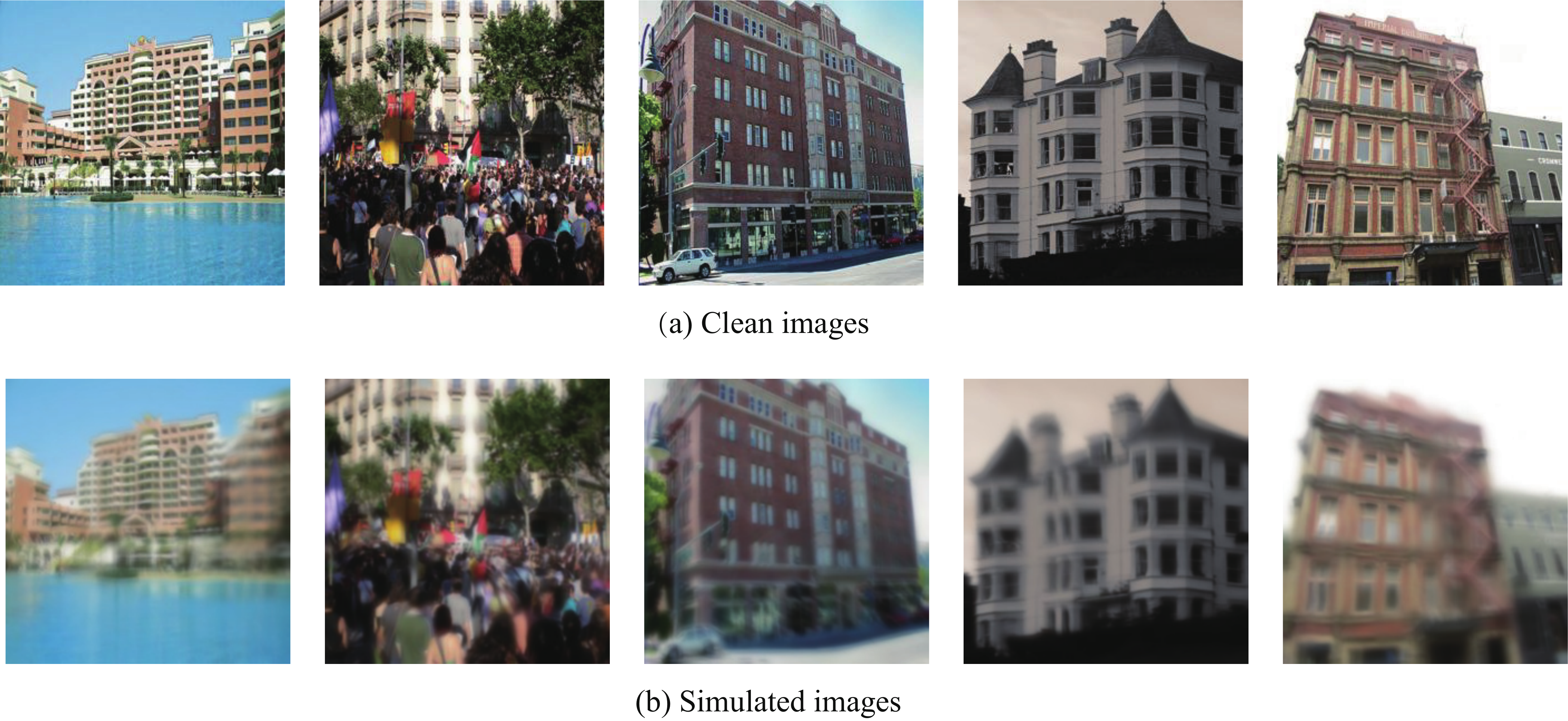}
  \caption{The visual examples of the training dataset. (a) shows clean images from place dataset \cite{zhou2017places} while (b) shows blurry images simulated by the state-of-the-art atmospheric turbulence simulator, P2S \cite{mao2021accelerating}.}
  \label{training}
\end{figure*}

\begin{table*}[htpb]
    \centering
    \begin{minipage}{\textwidth}
      \caption{The detailed configuration of the simulator}\label{configuration}
      \begin{tabular*}{\textwidth}{@{\extracolsep{\fill}}ccccccc@{\extracolsep{\fill}}}
      \toprule
      Stength & Probability & Kernel Size & \emph{D}(m)           & \emph{D}/$r_{0}$                 & Distance(m)  & Corr                     \\ \hline
      \midrule
      weak    & 0.5         & 33          & U(0.001,0.005) & {[}0.4,0.8,1.2,1.5{]} & U(150,600)   & {[}-1,-0.1,-0.5,-0.05{]} \\
      Medium  & 0.3         & 33          & U(0.04,0.1)    & {[}0.8,1,1.6{]}       & U(500,800)   & {[}-1,-0.1,-0.5,-0.05{]} \\
      Strong  & 0.2         & 33          & U(0.1,0.2)     & {[}1.6,2,2.4{]}       & U(1000,1500) & {[}-1,-0.1,-0.5,-0.05{]} \\ \hline
      \bottomrule  
      \end{tabular*}
    \end{minipage}
\end{table*}

\subsection{Singe Image Deblur}\label{sec2.3}
\begin{figure*}[htbp]
 \centering
    \includegraphics[width=1.00\linewidth]{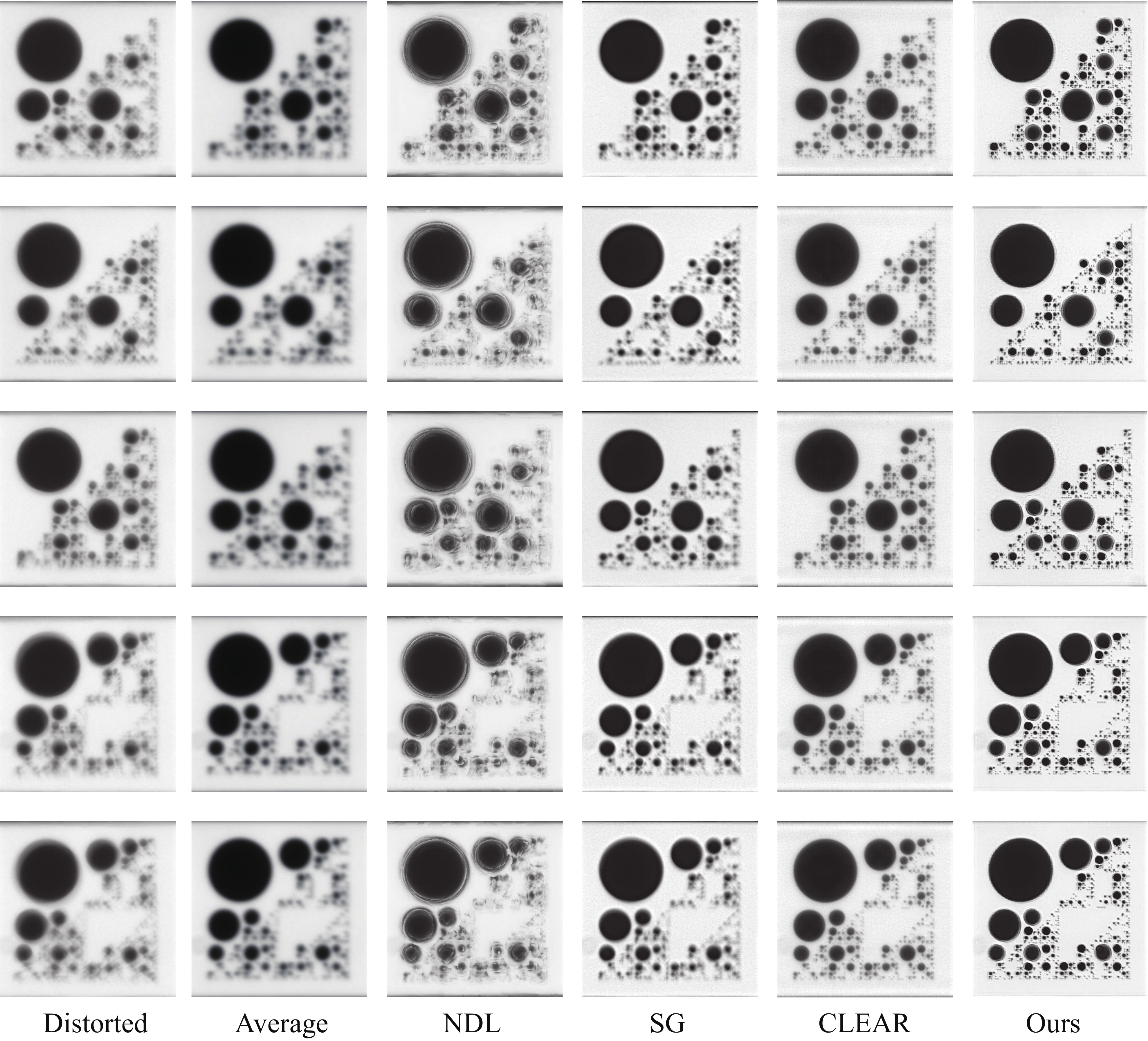}
 \caption{Visualization of reconstruction results on different kinds of code images.}
 \label{comparison}
\end{figure*}
Following the averaging of the selected frames, we apply a learning-based deblur method to remove any residual blur present in the averaged image \textbf{\textit{A}}, generating a sharp image \textbf{\textit{I}} without blurriness. To achieve this, we adopt the Uformer \cite{wang2022uformer} as our 
backbone, owing to its excellent performance in image restoration. Additionally, we generate our training dataset using an atmospheric turbulence simulator. A detailed description of the training dataset is provided in the subsequent section.
\subsection{Postprocess Refinement}\label{sec2.4}
Finally, post-processing techniques are employed to further enhance the quality of the output image \textbf{\textit{O}}, such as increasing its contrast and suppressing any ringing artifacts.
\section{Description of Training Dataset}\label{sec3}
Step C involves applying a learning-based deblurring method to remove any residual blur present in the averaged image A. As the training of deep neural networks requires a significant amount of data, we utilized the state-of-the-art P2S simulator \cite{mao2021accelerating} to synthesize a training dataset specifically for deblurring. Moreover, we exclusively used the simulator to simulate the blur caused by atmospheric turbulence, as the geometric distortion had already been corrected in the previous steps.
\subsection{Description of Ground Truth Dataset}\label{sec3.1}
In order to recover images that have been distorted by turbulence in diverse scenarios, 
the training dataset should include diverse scenes. Therefore, we utilized the place dataset \cite{zhou2017places}
for image synthesis. We selected 5000 images in the original dataset as input of the 
simulator. Fig. \ref{training}(a) and Fig. \ref{training}(b) are respectively clear images and simulated images.
\subsection{The Configuration of Simulator}\label{sec3.2}
In order to mitigate atmospheric turbulence of various intensities, we referenced the 
simulation method from \cite{zhang2022imaging} that achieved SOTA results in atmospheric turbulence mitigation. Specifically, we divide turbulence strength in 3 levels, they are weak,
medium and strong. The detailed configuration of the simulator is shown Table \ref{configuration}. From top to bottom the turbulence strength becomes higher. \textit{D} denotes the aperture diameter of the sensor, \emph{D}/$r_{0}$   denotes the ratio of \textit{D} and the fried parameter r0, Corr denotes the spatial correlation. U(a, b) denotes uniform distribution in the range(a, b) and [] 
denotes random choice with equal probability.
\section{Experiments}\label{sec4}

\subsection{Datasets and Experimental Setting}\label{sec4.1}
\noindent\textbf{Final Test Datasets.} We conduct experiments on 192 kinds of code sequences to evaluate the proposed framework, each suquence is composed of 100 distorted frames.

\noindent\textbf{Experimental Setting.} We compare the proposed framework with some existing atmospheric turbulence mitigation methods, including CLEAR \cite{anantrasirichai2013atmospheric}, SG \cite{lou2013video}, NDL \cite{zhu2012removing}.

\subsection{Experiments on Code Images}\label{sec4.2}
We evaluate the performance of the proposed framework and some existing methods on the final test datasets. The visual results of different methods are shown in Fig. \ref{comparison}.
As for visual comparison in the results, it is evident that NDL \cite{zhu2012removing} struggle to handle the datasets and leave the geometric distortion uncorrected. Although CLEAR \cite{anantrasirichai2013atmospheric} and SG \cite{lou2013video} can effectively mitigate geometric distortion, some degree of residual blurriness may still be present in the output images. 
Compared with these methods, our proposed framework demonstrates superior performance, as it can simultaneously mitigate both blur and geometric distortion.

\section{Conclusion}
In our submission to the track 2.2 in UG$^{2}$+ Challenge in CVPR 2023, we propose an efficient multi-stage framework to restore a high-quality image from distorted frames. Firstly, A frame selection algorithm based on sharpness is first utilized to select the best set of distorted frames.
is aligned using optical-flow based registration to suppress
geometric distortion. Next, we select the sharpest set of
registered frames using frame selection based on sharpness
and average them to produce an image approximately free
of geometric distortion but with blurriness. Then, a learning-
based deblur method is applied to remove the blur in the
average image. Finally, post-processing is utilized to further
enhance the quality of the output image. Our framework can handle different kinds of code dataset provided in the final testing phase, 
and ranked 1st on the final leaderboard. In the future, we will
explore more efficient methods to improve this task.
{\small
\bibliographystyle{plain}
\bibliography{refpaper.bib}
}

\end{document}